\pdfoutput=1

\documentclass[11pt]{article}

\usepackage[]{acl}

\usepackage{times}
\usepackage{latexsym}

\usepackage[T1]{fontenc}

\usepackage[utf8]{inputenc}

\usepackage{microtype}

\usepackage{amsmath}

\usepackage{graphicx}
\usepackage{xcolor}
\usepackage{multirow}
\usepackage{multicol}
\usepackage{centernot}
\usepackage{tabularx}
\usepackage{pgfplots}
\usepackage{pifont}
\newcommand{\cmark}{\ding{51}}%
\newcommand{\xmark}{\ding{55}}%
\usepackage{tikz}
\usepackage{cleveref}

\crefformat{section}{\S#2#1#3}
\crefformat{subsection}{\S#2#1#3}
\crefformat{subsubsection}{\S#2#1#3}

\usetikzlibrary{patterns}
\pgfplotsset{compat=newest}
\pgfplotsset{every axis/.append style={line width=1pt
}}
\definecolor{color1}{HTML}{601A4A}
\definecolor{color2}{HTML}{EE442F}
\definecolor{color3}{HTML}{63ACBE}

\definecolor{mygreen}{RGB}{29, 177, 0}

\newcommand{\minisection}[1]{\noindent{\bf #1}\hspace{0.6em}}
\newcommand{\mediumsection}[1]{\vspace{0.5em}\noindent{\bf #1}\hspace{0.6em}}

\newcommand{\notimplies}{\centernot\implies}

\title{Is ``\textit{My Favorite New Movie}'' My Favorite Movie?\\ Probing the Understanding of Recursive Noun Phrases}

\author{\textbf{Qing Lyu}$^1$\hspace{5mm}\textbf{Hua Zheng}$^2$ \hspace{5mm}\textbf{Daoxin Li}$^3$ \hspace{5mm} \textbf{Li Zhang}$^1$ \\\textbf{Marianna Apidianaki}$^1$ \hspace{5mm} \textbf{Chris Callison-Burch}$^1$\\
$^1$Department of Computer and Information Science, University of Pennsylvania\\
$^2$Key Lab of Computational Linguistics (MOE), Peking University\\
$^3$Department of Linguistics, University of Pennsylvania\\
\texttt{\{lyuqing,zharry,marapi,ccb\}@seas.upenn.edu}\\
\texttt{zhenghua@pku.edu.cn}\\
\texttt{daoxinli@sas.upenn.edu}\\
}

\begin{document}
\maketitle
\begin{abstract}
Recursive noun phrases (NPs) have interesting semantic properties. For example, \textit{my favorite new movie} is not necessarily my favorite movie, whereas \textit{my new favorite movie} is. This is common sense to humans, yet it is unknown whether language models have such knowledge. We introduce the Recursive Noun Phrase Challenge (\textbf{RNPC}), a dataset of three textual inference tasks involving textual entailment and event plausibility comparison, precisely targeting the understanding of recursive NPs. When evaluated on RNPC, state-of-the-art Transformer models only perform around chance. Still, we show that such knowledge is learnable with appropriate data. We further probe the models for relevant linguistic features that can be learned from our tasks, including modifier semantic category and modifier scope. Finally, models trained on RNPC achieve strong zero-shot performance on an extrinsic Harm Detection evaluation task, showing the usefulness of the understanding of recursive NPs in downstream applications.\footnote{Our code and data are available at \url{https://github.com/veronica320/Recursive-NPs}.}

\end{abstract}

\section{Introduction}



\begin{figure}[t!]
\includegraphics[width=\columnwidth]{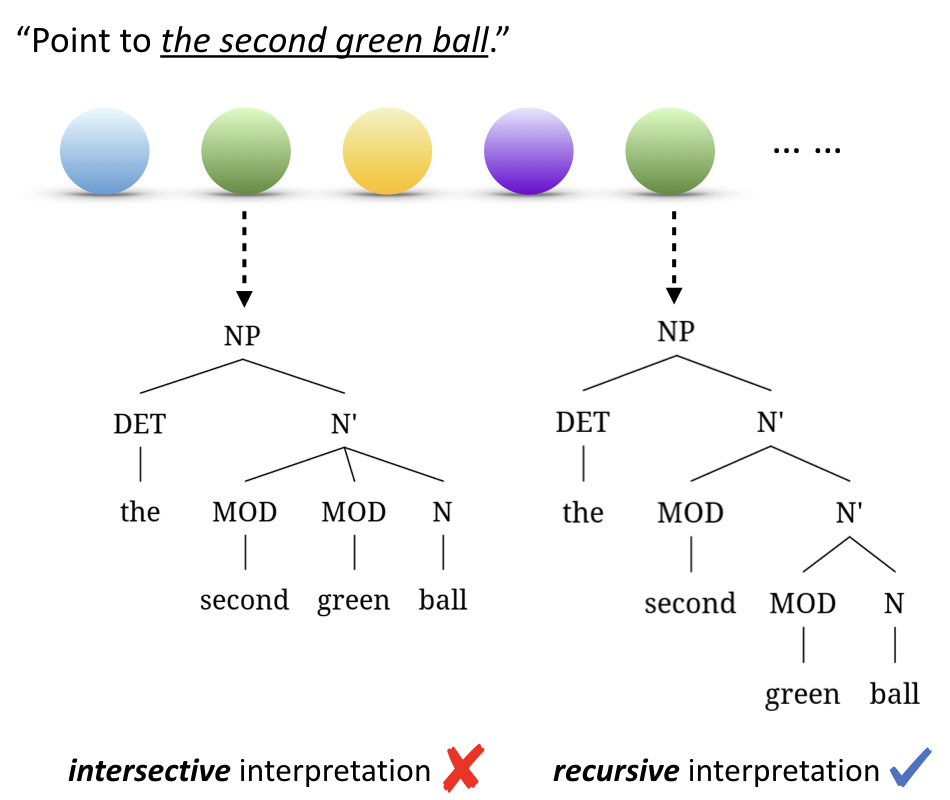}
\caption{The intersective (incorrect) and the recursive (correct) interpretation of \textit{the second green ball}.}\label{figure:thumbnail}
\vspace{-0.2in}
\end{figure}

Recursion, the self-embedding of a linguistic structure, constitutes a fundamental property of human language. Due to its hierarchical structure, it poses many challenges to human language acquisition. One such challenge occurs in the context of recursive Noun Phrases (NPs), i.e., NPs with multiple prenominal modifiers. For instance, in Figure~\ref{figure:thumbnail}, when asked to point to \textit{the second green ball} in a series of balls, children sometimes erroneously point to the second \textbf{and} green ball (intersective interpretation), instead of the second \textbf{among} green balls (recursive interpretation) \cite{matthei1982acquisition, hamburger1984acquisition, marcilese2013recursive}.

We investigate whether language models (LMs) make similar errors, since the understanding of recursive NPs is also fundamental in real-world AI applications. For example, a summarization system should know that \textit{the former US president} cannot be shortened as \textit{the president}, since they are no longer in power. Also, a self-driving car asked to take \textit{the first left-hand exit} should not assume that it is always the first exit.

\begin{table*}[!t]
\centering
\scalebox{0.8}{
\begin{tabular}{lcll}
    \hline \textbf{Task} & \textbf{ID} & \textbf{Input} & \textbf{Label} \\
    \hline \multirow{4}{3cm}{Single-Premise Textual Entailment (SPTE)} & \multirow{2}{*}{(1a)} &Premise: This is \underline{my \textbf{new favorite} movie}.   & \multirow{2}{2cm}{Entailment}\\
    & & Hypothesis: This is \underline{my favorite movie}. & \vspace{0.025in}\\
    & \multirow{2}{*}{(1b)} &Premise: This is \underline{my \textbf{favorite new} movie}. & \multirow{2}{*}{Non-Entailment}\\
    & & Hypothesis: This is \underline{my favorite movie}. & \vspace{0.05in}\\
    
    \hline \multirow{6}{3cm}{Multi-Premise Textual Entailment (MPTE)} & \multirow{3}{*}{(2a)} &Premise 1: He is \underline{a skillful American violinist}.   & \multirow{3}{*}{Entailment}\\
    & & Premise 2: He is \underline{a father}. & \\
    & & Hypothesis: He is \underline{an \textbf{American} father}. & \vspace{0.025in}\\
    & \multirow{3}{*}{(2b)} &Premise 1: He is \underline{a skillful American violinist}.   & \multirow{3}{*}{Non-Entailment}\\
    & & Premise 2: He is \underline{a father}. & \\
    & & Hypothesis: He is \underline{a \textbf{skillful} father}. & \vspace{0.05in}\\
    
    \hline \multirow{6}{3cm}{Event Plausibility Comparison (EPC)} & \multirow{2}{*}{(3a)} &Event 1: \underline{The actress} is known by everyone. & \multirow{2}{2.5cm}{(Event 2 is) More Plausible}\\
    & & Event 2: \underline{The famous former actress} is known by everyone. & \vspace{0.025in}\\
    
    & \multirow{2}{*}{(3b)}& Event 1: \underline{The actress} lives in France. & \multirow{2}{2.7cm}{(Event 2 is) Equally Plausible}\\
    & & Event 2: \underline{The famous former actress} lives in France. & \vspace{0.025in}\\
    
    & \multirow{2}{*}{(3c)}& Event 1: \underline{The actress} stars in many latest movies. & \multirow{2}{2.5cm}{(Event 2 is) Less Plausible}\\
    & & Event 2: \underline{The famous former actress} stars in many latest movies. & \vspace{0.025in}\\ 
    \hline
\end{tabular}
}
 \vspace{-0.05in}
\caption{Examples for each task in our dataset. The NPs of interest are underlined. Differences between examples are in bold. See Section~\ref{sec:formulation} for details.}\label{table:task_examples}
 \vspace{-0.1in}
\end{table*}

Previous work has studied the syntactic parsing of recursive NPs \cite{nakov-hearst-2005-search, pitler-etal-2010-using}, as well as the semantic categorization of modifiers in NPs with only one prenominal modifier \cite{kamp1995prototype, mccrae-etal-2014-modelling}. 
However, neither parsing nor modifier categorization alone can sufficiently capture the meaning of recursive NPs (\cref{sec:related_work}).


In this paper, using recursive NPs with two modifiers as our test-bed, we address the following questions about LMs' understanding of recursion:

(a) \textbf{Is the knowledge of how to interpret recursive NPs present in LMs (\cref{section:Qa})?} We propose the Recursive Noun Phrase Challenge (RNPC), a challenge set containing three classification tasks: Single-Premise Textual Entailment, Multi-Premise Textual Entailment, and Event Plausibility Comparison (\cref{sec:formulation}). Table~\ref{table:task_examples} provides examples for each task. Results show that state-of-the-art (SOTA) LMs finetuned on standard benchmarks of the same format (e.g., MNLI \cite{williams2018broad}) all struggle on our dataset, suggesting that the target knowledge is not readily available.

(b) \textbf{Is such knowledge learnable with appropriate data (\cref{section:Qb})?} We adopt the challenge set analysis technique proposed by \citet{liu-etal-2019-inoculation}, which exposes models to a small amount of data and assesses how well they can adapt. All models achieve a noticeable performance improvement with as few as 200 examples, indicating that the target knowledge is potentially learnable.

(c) \textbf{What can models learn from recursive NPs (\cref{section:Qc})?} We probe the finetuned models for two well-studied linguistic features in previous work, modifier semantic category and modifier scope. We show that both features can be learned from RNPC, with techniques including edge probing \cite{tenney2018what} and attention visualization \cite{vig-2019-multiscale}.

(d) \textbf{Is such knowledge useful for downstream tasks (\cref{section:Qd})?} When evaluated on an extrinsic Harm Detection task, models finetuned on RNPC achieve strong zero-shot performance. This shows that the understanding of recursive NPs can benefit downstream language understanding tasks.

In summary, our work identifies an interesting linguistic phenomenon that is common sense to humans but challenging for models. It contributes to the characterization of LMs' limitations and capabilities in language understanding.






\section{Related Work}
\label{sec:related_work}

Noun Phrases (NPs) have been extensively studied in both linguistics and NLP, primarily from the following perspectives.

\minisection{Syntactic structure.} A line of work focuses on the syntactic structure of NPs, which essentially explains the \textbf{modifier scope} \cite{campbell-2002-computation} in NPs. One classic task is NP bracketing, i.e., deciding whether an NP is right-branching (e.g., [\textit{world} [\textit{oil prices}]]) or left-branching (e.g., [[\textit{crude oil}] \textit{prices}]) \cite{lauer1995corpus, nakov-hearst-2005-search}. A harder task is full parsing \cite{ vadas-curran-2007-adding, pitler-etal-2010-using}, i.e., reconstructing the complete dependency tree.

\minisection{Modifier semantics.} Another line of research revolves around the semantics of simple modifier-noun composition, starting with ways to \textbf{categorize modifiers} based on their inference patterns \cite{kamp1995prototype, bouillon1999description, chierchia2000meaning}. With $M$ as the modifier and $N$ as the noun, a representative taxonomy summarized by \citet{mccrae-etal-2014-modelling} is:\\
(1) {\bf intersective}: 
$X$ is a $M$ $N$ $\implies X$ is $M \land X$ is a $N$, e.g., ``an \textit{American} surgeon'' describes someone who is both American and a surgeon;\\
(2) {\bf subsective}: $X$ is a $M$ $N$ $\implies X$ is a $N$, but $X$ is a $M$ $N$ $\notimplies X$ is $M$, e.g., someone who is ``a \textit{skillful} surgeon'' is not necessarily skillful in all disciplines;\\
(3) {\bf privative}: $X$ is a $M$ $N$ $\notimplies X$ is a $N$, e.g., ``a \textit{former} surgeon'' describes someone who is no longer a surgeon.

Despite the variations\footnote{For example, other studies call category (3) ``non-subsective'' instead, and further decompose it into ``privative'' ($X$ is a $M$ $N$ contradicts $X$ is a $N$, e.g., \textit{fake}) and ``non-privative'' ($X$ is a $M$ $N$ is neutral to $X$ is a $N$, e.g., \textit{alleged}).} and debates\footnote{Some linguists (for example, \citet{partee2010privative}) argue that (3) should be subsumed by (2), since privative modifiers can coerce the noun they modify into a looser interpretation.} on the taxonomy, we follow these conventional terms in subsequent sections.


With the advances in NLP, more recent works starts modeling the semantics of simple modifier-noun constructions with first-order logic \cite{mccrae-etal-2014-modelling}, linear mapping \cite{baroni-zamparelli-2010-nouns}, and other explicit compositional operations \cite{boleda-etal-2012-first, boleda-etal-2013-intensionality}. In particular, \citet{pavlick-callison-burch-2016-babies, pavlick-callison-burch-2016-called} propose a novel contextualized inference-based approach. They define the Add-One Entailment task with natural contexts from textual corpora, where the hypothesis differs from the premise by the insertion of one modifier. For example, \textit{The \underline{crowd} roared} entails \textit{The \underline{enthusiastic crowd} roared}, though \textit{enthusiastic crowd} denotes a subset of \textit{crowd} without context. However, natural contexts also introduce complications from monotonicity \cite{van1983determiners}. For instance, \textit{red apple} entails \textit{apple}, but \textit{He didn't eat any red apple} does not entail \textit{He didn't eat any apple} due to the downward entailment context. In our proposed approach, we handle this issue by controlling for context monotonicity.

Other related work explores which attributes of the head noun are affected by the presence of modifiers. \citet{mullenbach2019nuclear} look at how modifiers project from a noun to its parts (e.g., does a \textit{red jeep} have \textit{red tires}?). \citet{emami-etal-2021-adept} test the likelihood change of an event when a modifier is added (e.g., \textit{a false key} is less likely to \textit{open a door} than \textit{a key}). \citet{apidianaki-gari-soler-2021-dolphins} study the prototypical properties of nouns (e.g., \textit{a strawberry} entails \textit{a red strawberry}). Researchers also examine the interpretation of noun compounds \citep{shwartz-waterson-2018-olive, hendrickx-etal-2013-semeval} (e.g., olive oil is made \textit{of} olives, while baby oil is made \textit{for} babies).
 


\minisection{Summary.} Neither syntactic parsing nor modifier semantics alone can fully capture the meaning of recursive NPs. In terms of syntax, modifier scope cannot always explain NPs due to the influence from modifier semantics. For instance, \textit{a [big [fake gun]]} and \textit{a [big [black gun]]} have the same structure but different inference patterns, i.e. only the latter is a gun. 
Meanwhile, modifier category itself does not suffice without taking into account modifier scope. For example, \textit{a so-called healthy food} and \textit{a so-called homeopathy expert} start with the same privative modifier (\textit{so-called}). However, \textit{so-called} questions truthfulness of the second modifier (\textit{healthy}) in the former case while that of the noun (\textit{expert}) in the latter. Therefore, we introduce a dataset containing three novel and challenging textual inference tasks, which rely on the interplay of syntax and semantics in determining the meaning of recursive NPs. 

\vspace{-0.03in}
\section{Task Formulation}
\label{sec:formulation}
\vspace{-0.03in}

Our dataset contains three tasks. Let us denote a canonical two-modifier recursive NP by $\mathbf{Det\ M_1\ M_2\ N}$ (Determiner, Modifier 1, Modifier 2, Noun). With this notation, the tasks are outlined below. See Table~\ref{table:task_examples} for concrete examples.

\mediumsection{Single-Premise Textual Entailment (SPTE)} follows the conventional TE task format. Given a premise and a hypothesis, the model decides whether the premise semantically entails the hypothesis.
The labels include \texttt{entailment} and \texttt{non-entailment}.\footnote{We do not distinguish between \texttt{neutral} and \texttt{contradiction} in order to minimize label ambiguity.} An SPTE example can be represented in regular expression as:
\begin{align*}
\textbf{Premise}&: P\ Det\ M_1\ M_2\ N\\
\textbf{Hypothesis}&: P\ Det\ (M_1|M_2)?\ N\\
\textbf{Label}&: \texttt{entailment}|\texttt{non-entailment}
\end{align*}
where $\mathbf{P}$ is a sentence prefix, which can be instantiated as \textit{This is/He is/She is}, etc., depending on the NP. Intuitively, this task tests \textbf{whether an NP entails its various components}. This holds for most simple NPs (e.g., \textit{the second ball} entails \textit{ball}), but recursive NPs offer interesting counterexamples (e.g., (1b) in Table~\ref{table:task_examples}).

\mediumsection{Multi-Premise Textual Entailment (MPTE)} is adapted from the attributive propagation test described in \citet{lalissedistinguishing}. The format differs from SPTE only in that it has two premises instead of one. Given that both are true, the task is to determine whether the hypothesis is also true. The first premise is of the same form as in SPTE. The second premise contains a noun other than $\mathbf{N}$, denoted by $\mathbf{N_2}$.\footnote{For both premises to hold at the same time, we need an $\mathbf{N_2}$ that can refer to the same entity as $\mathbf{N}$.} A regular expression representation is:
\begin{align*}
\textbf{Premise 1}&: P\ Det\ M_1\ M_2\ N\\
\textbf{Premise 2}&: P\ Det\ N_2\\
\textbf{Hypothesis}&: P\ Det\ (M_1|M_2)\ N_2\\
\textbf{Label}&: \texttt{entailment}|\texttt{non-entailment}
\end{align*}
This test targets the \textbf{compositionality of modifiers and nouns}. While most of the time a modifier can be freely ``detached'' and ``attached'' (e.g., (2a)), sometimes it cannot (e.g., (2b)).

\mediumsection{Event Plausibility Comparison (EPC)} follows the task formalization by \citet{emami-etal-2021-adept} for single-modifier NPs. Given two events, $\mathbf{Event 1}$ and $\mathbf{Event 2}$, a model needs to assess the plausibility of $\mathbf{Event 2}$ compared to that of $\mathbf{Event 1}$. The two events have the same event predicate $\mathbf{E}$, and differ only in the $\mathbf{NP}$. A regular expression representation is:
\begin{align*}
\textbf{Event 1}&: Det\ (M_1|M_2)?\ N\ E\\
\textbf{Event 2}&: Det\ M_1\ M_2\ N\ E\\
\textbf{Label}&: \texttt{more}|\texttt{equally}|\texttt{less plausible}
\end{align*}
This task tests the \textbf{influence of adding modifier(s) on the plausibility of different events about the noun}. Not all events are affected in the same way: in (3), \textit{stars in many latest movies} becomes less plausible, while \textit{is known by everyone} is more so.

\begin{table}[!t]
\begin{center}
\scalebox{0.8}{
\begin{tabular}{p{1.8cm}|p{1cm}p{5.5cm}}
\hline \bf Category & \bf Count & \textbf{Examples}: modifier (\textsc{attribute}) \\ 
\hline Intersective & 296 & red (\textsc{color}), female (\textsc{gender}), \newline German (\textsc{nationality})\\
Subsective & 269 & short (\textsc{height}), small (\textsc{size}), \newline far (\textsc{distance}) \\
Privative & 124 & former (\textsc{time}), vice (\textsc{authority}), fake (\textsc{authenticity}) \\
\hline
\end{tabular}
}
\end{center}
\caption{Statistics and examples for each semantic category in our modifier lexicon.}\label{table:lexicon}
\end{table}

We choose the three tasks defined above because they allow us to study different interesting properties of recursive NPs that conventional parsing tasks do not. For example, SPTE is convenient for comparing the impact of modifier order on the meaning of the NP (e.g., (1a) and (1b)); MPTE precisely reflects the property of subsective modifiers (e.g., \textit{skillful}); whereas EPC is suitable for NPs with privative modifiers, since the other formats often cause ambiguity in this case.\footnote{For example, \textit{fake fur} might or might not be considered a kind of fur \cite{partee2010privative}. Annotators would thus probably disagree on the label if it were an SPTE example.}

\section{Dataset Construction}

Our dataset is constructed in four stages: (a) modifier lexicon construction, (b) NP extraction and selection, (c) instance creation and review, and (d) label verification. Among them, (c) and (d) involve crowdsourcing.\footnote{See more statistics, crowdsourcing setup, and agreement details in Appendix~\ref{appendix:data_construction}; see annotation guidelines and HIT design in the Supplementary Materials.}

\begin{table}[!t]
\begin{center}
\scalebox{0.9}{
\begin{tabular}{p{1cm}|p{1cm}p{1.3cm}p{1.7cm}}
\hline Task & \bf Total & Entail & Non-entail \\ 
\hline SPTE & 1,163 & 582 & 581 \\
\hline
\end{tabular}
 \vspace{0.05in}
}
\scalebox{0.9}{
\begin{tabular}{p{1cm}|p{1cm}p{1.3cm}p{1.7cm}}
\hline Task & \bf Total & Entail & Non-entail \\ 
\hline MPTE & 1,063 & 541 & 522 \\
\hline
\end{tabular}
}
\vspace{0.05in}
\scalebox{0.9}{
\begin{tabular}{p{1cm}|p{1cm}p{0.85cm}p{0.85cm}p{0.85cm}}
\hline Task  & \bf Total  & More & Equal &  Less \\ 
\hline EPC & 1,479 & 508 & 392 & 579 \\
\hline
\end{tabular}
}
\end{center}
 \vspace{-0.15in}
\caption{Number of examples in each RNPC task. Entail/Non-entail stand for Entailment/Non-entailment, and More/Equal/Less stand for More Plausible/Equally Plausible/Less Plausible.}\label{table:stats}
\end{table}

\minisection{Modifier lexicon construction.}
We first construct a lexicon of modifiers following the taxonomy in Section~\ref{sec:related_work} \citep{mccrae-etal-2014-modelling}. We include modifiers studied in relevant linguistics literature \cite{nayak2014dictionary,lalissedistinguishing} and complement the list with modifiers that are missing or have not been addressed before under this lens (for example, modifiers that describe material, such as \textit{wooden}, can also be viewed as privative). Each entry in the lexicon contains the modifier itself, its category (intersective, subsective, or privative), and its attribute (e.g., \textit{green} is a \textsc{color}). In total, the lexicon contains 689 modifiers, the largest resource of this kind. See Table~\ref{table:lexicon} for category distribution and examples.

\begin{figure*}[t!]

\includegraphics[width=\textwidth]{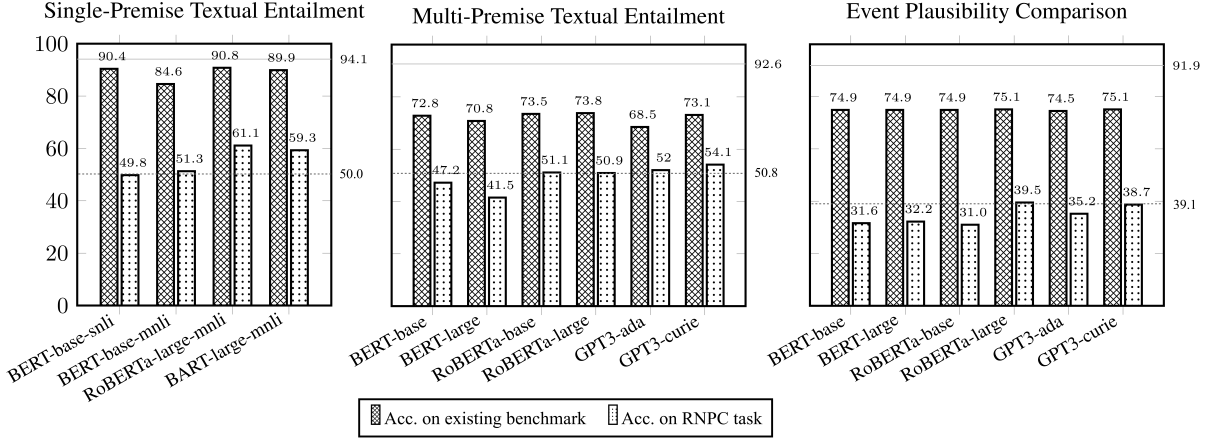}
\centering
\vspace{-0.2in}
\caption{Given SOTA models finetuned on existing benchmark(s) of the same format as each RNPC task, we compare their accuracy on these benchmark(s) and on the RNPC task. The dotted line represents the majority baseline, and the solid line stands for human performance. Models for SPTE are finetuned on MNLI and SNLI, while models for the other two tasks are finetuned on MPE and ADEPT, respectively.}\label{figure:eval_perf}
\end{figure*}

\minisection{NP extraction and selection.} 
Next, we collect recursive NPs from a variety of resources: linguistics literature \cite{matthei1982acquisition, 10.1007/11424918_35, teodorescu2006adjective, morzycki2016modification}, text corpora (Penn Treebank \cite{marcus1993building} and the Annotated Gigaword corpus \cite{napoles2012annotated}), and our creation. From text corpora, we extract all NPs with more than two consecutive modifiers in our lexicon, and manually select NPs considering a set of factors: lexical diversity, class balance, whether there is an interaction between the modifiers, etc. Finally, we complement the set with deliberately designed challenging cases of our invention, resulting in 1,299 NPs in total.

\minisection{Instance creation and review.} We hire college students\footnote{Specifically, undergraduate and graduate students in an Artificial Intelligence class.} to write examples for the three tasks based on our collection of NPs. Each student is given a screening test containing five NPs. If $\geq 75\%$ of their created examples across all tasks are valid, they are qualified to continue. Each instance is then reviewed and/or revised by one of the authors, resulting in 8,260 valid instances.

\minisection{Label verification.} We again hire college students to verify instance labels via Amazon Mechanical Turk. Each task has a screening test of 10 easy instances with an unambiguous answer, and only students with an accuracy of $\geq90\%$ can proceed. During the official annotation, a HIT contains 10 questions of a task, including one control question. Each HIT is completed by three people, excluding its creator. Annotations are then filtered based on the accuracy on control questions and the time used. Only examples with $\geq2$ people agreeing with the gold label are retained, yielding 4,567 examples. We then down-sample the examples in each task for a relatively balanced ratio among classes, resulting in 3,705 examples. See Table~\ref{table:stats} for details.

\section{Do LMs understand recursive NPs?}
\label{section:Qa}

To answer question (a), whether the knowledge of how to interpret recursive NPs is present in pretrained LMs, we use the ``behavioral test'' probing method \cite{belinkov-etal-2020-interpretability}. Namely, we evaluate SOTA models finetuned on existing benchmark(s) of the same format as each RNPC task. The rationale is that LMs should acquire the ability of textual inference in the required format during finetuning, which allows us to elicit their potential knowledge about recursive NPs.\footnote{LMs can also overfit the finetuning dataset and thus ``forget'' the target knowledge acquired during pretraining. Thus, we also directly probe the pretrained LMs in a complementary ``likelihood scoring'' experiment, described in Appendix~\ref{appendix:direct_probe}.}


\minisection{Experimental setup.} We consider the following datasets that address similar phenomena as our tasks: (1) MNLI \cite{williams2018broad} and SNLI \cite{bowman-etal-2015-large} for our SPTE; (2) MPE \cite{lai-etal-2017-natural} for our MPTE; and (3) ADEPT \cite{emami-etal-2021-adept} for our EPC. We choose SOTA and close-to-SOTA models on these benchmarks as probing candidates, including BERT \cite{devlin-etal-2019-bert}, RoBERTa \cite{liu2019roberta}, BART \cite{lewis-etal-2020-bart}, and GPT3 \cite{NEURIPS2020_1457c0d6}.\footnote{Due to the size of MNLI and SNLI, we only evaluate available checkpoints from the Huggingface Transformers model hub. For the other two benchmarks, all models are trained by us. Also, the largest GPT3-davinci is unavailable for finetuning and thus excluded. See Appendices~\ref{appendix:benchmark_details} and ~\ref{appendix:implementation_eval_model} for dataset, model and hyperparameter details.}

\begin{table*}[!t]
\centering
\scalebox{0.78}{
\begin{tabular}{lclll}
    \hline \textbf{Task} & \textbf{ID} & \textbf{Input} & \textbf{Gold Label} & \textbf{Predicted Label}\\
    \hline \multirow{4}{2cm}{Single-Premise Textual Entailment} & \multirow{2}{*}{(1a)} &Premise: This is my \underline{new favorite} movie.   & \multirow{2}{3cm}{Entailment} & \multirow{2}{3cm}{\textcolor{blue}{Entailment \cmark}}\\
    & & Hypothesis: This is my favorite movie. & \vspace{0.025in}\\
    & \multirow{2}{*}{(1b)} &Premise: This is my \underline{favorite new} movie. & \multirow{2}{3cm}{Non-Entailment} & \multirow{2}{3cm}{\textcolor{red}{Entailment \xmark}} \\
    & & Hypothesis: This is my favorite movie. & \vspace{0.05in}\\
    
    \hline \multirow{6}{2cm}{Multi-Premise Textual Entailment} & \multirow{3}{*}{(2a)} &Premise 1: He is a short American basketball player.   & \multirow{3}{3cm}{Entailment} & \multirow{3}{3cm}{\textcolor{blue}{Entailment \cmark}} \\
    & & Premise 2: He is a man. & \\
    & & Hypothesis: He is an \underline{American} man. & \vspace{0.025in}\\
    & \multirow{3}{*}{(2b)} &Premise 1: He is a short American basketball player.   & \multirow{3}{3cm}{Non-Entailment} & \multirow{3}{3cm}{\textcolor{red}{Entailment \xmark}}\\
    & & Premise 2: He is a man. & \\
    & & Hypothesis: He is a \underline{short} man. & \vspace{0.05in}\\
    
    \hline \multirow{4}{2cm}{Event Plausibility Comparison} & \multirow{2}{*}{(3a)} &Event 1: An animal can be harmful to people. & \multirow{2}{3cm}{Less Plausible} & \multirow{2}{3cm}{\textcolor{blue}{Less Plausible \cmark}} \\
    & & Event 2: A \underline{dead dangerous} animal can be harmful to people. & \vspace{0.025in}\\
    
    & \multirow{2}{*}{(3b)}& Event 1: An animal can be harmful to people. & \multirow{2}{3cm}{More Plausible} & \multirow{2}{3cm}{\textcolor{red}{Less Plausible \xmark}} \\
    & & Event 2: A \underline{dangerous dead} animal can be harmful to people. & \vspace{0.025in}\\
    
    \hline
\end{tabular}
}
 \vspace{-0.05in}
\caption{Minimal-pair examples where the best-performing models make errors for each RNPC task. Differences between each pair are underlined. }\label{table:minimal_pair}
\end{table*}

\minisection{Results and analysis.} 
We evaluate the finetuned models on each RNPC task. When the finetuning dataset has more classes than our task does, we map the model prediction to one of our classes by summing probability scores.\footnote{For example, for a model trained on MNLI (with three labels), we compare the score of \texttt{entailment} and the summed score of \texttt{neutral} and \texttt{contradiction}. If the former is higher, we predict \texttt{entailment} on SPTE; otherwise \texttt{non-entailment}. Empirically, this strategy results in higher performance than directly mapping the highest-score MNLI label to its corresponding SPTE label.} Figure~\ref{figure:eval_perf} compares the performance of the models on the relevant benchmarks and our tasks. We also include human performance, calculated by averaging the accuracy of three college student annotators on a random sample of 300 examples for each task.

All models struggle on RNPC with performance around chance, while human accuracy is constantly above 90. On SPTE and MPTE, almost all models have a high false-positive rate. As long as all tokens in the hypothesis (e.g., \textit{This is the second ball}) appear in the premise (e.g., \textit{This is the second green ball}), they tend to predict \texttt{entailment}, indicating that they are making the same intersective interpretation errors as children do. On EPC, most models over-predict \texttt{equally plausible}, arguably due to the class imbalance during finetuning. This also shows that our task is not trivially solvable by models that understand non-recursive NPs, which the finetuning dataset comprises.

Next, we closely examine the best-performing models on each task, including RoBERTa-large finetuned on MNLI, GPT3-curie finetuned on MPE, and RoBERTA-large finetuned on ADEPT. On MPTE and EPC, even the best model barely surpasses chance performance. On SPTE, the best accuracy (61.2) is still unimpressive for a binary classification task.
To understand where exactly the models fail, we further present a qualitative minimal-pair analysis in Table~\ref{table:minimal_pair}. On SPTE, the two examples differ only in the order of modifiers (\textit{new} and \textit{favorite}) in the premise, leading to opposite labels. However, the model predicts \texttt{entailment} for both, suggesting its insensitivity to subtle meaning differences incurred by modifier order changes. On MPTE, the difference between the two examples lies in the modifier in the hypothesis, \textit{an American man} vs. \textit{a short man}. As basketball players are generally tall, the second hypothesis should not be entailed. Again, the model predicts \texttt{entailment} for both cases, which shows its lack of relevant world knowledge. Finally, on EPC, \textit{a dead dangerous animal} and \textit{a dangerous dead animal} have subtly different meanings -- the former refers to a \textit{dangerous animal} that is dead (e.g., a dead lion, which is no longer harmful to people), while the latter refers to a \textit{dead animal} that has become dangerous (e.g., a dead squirrel carrying viruses, which is indeed harmful). The model fails to distinguish between them, predicting \texttt{less plausible} for both. All the above observations show that the knowledge for interpreting recursive NPs is not present in LM representations.

\vspace{-0.05in}
\section{Can LMs Learn the Meaning of Recursive NPs?}
\label{section:Qb}
\vspace{-0.05in}

We investigate the reasons behind the models' low performance on RNPC, specifically whether their failure is due to the lack of in-domain training data or an intrinsic deficiency in their architecture. Namely, we attempt to answer question (b): Is the target knowledge learnable with appropriate data?

We adopt the challenge set analysis technique from \citet{liu-etal-2019-inoculation}, which exposes a model to a small amount of challenge data and assesses how well it can adapt. Specifically, we split each RNPC task dataset into a training set of 200 examples and a new test set containing the rest, ensuring that they have different modifiers in the same position. For example, if a modifier appears as the $M_1$ of an NP in the training set, it cannot appear in the same position of any NP in the test set. Then, we finetune each model from Figure~\ref{figure:eval_perf} on an increasing number of examples (10 to 200). The learning curves of the best-performing models (RoBERTa-large (MNLI), RoBERTa-base (MPE), and RoBERTA-large (ADEPT)) are plotted in Figure~\ref{figure:inoculation_results}.\footnote{See Appendix~\ref{appendix:implementation_ino_model} for model and hyperparameter details.}

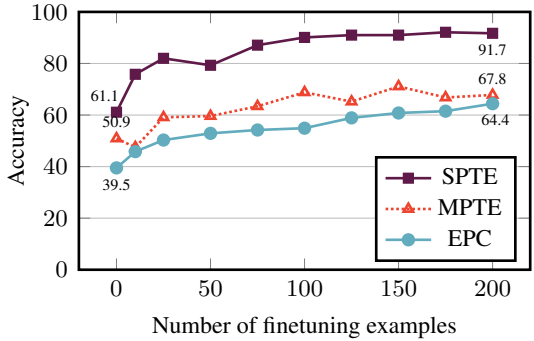
\begin{figure}[!t]
\centering
\begin{tikzpicture}
\begin{axis}[legend pos=south east,
    ymajorgrids,
    width=0.47\textwidth,
    height=5cm,
	ymin=0,
	ymax=100,
	ylabel = {Accuracy},
	y label style={font=\small, yshift=-0.25cm},
	y tick label style={font=\small},
	ytick distance=20,
	xlabel = {Number of finetuning examples},
	x label style={font=\small},
	x tick label style={font=\small},
	legend style={legend columns=1, font=\small}
]
\addplot[color=color1,mark=square*,mark options={scale=0.8}] coordinates {(0, 61.1) (10, 75.8) (25, 82.0) (50, 79.3) (75, 87.1) (100, 90.1) (125, 91.0) (150, 91.0) (175, 92.1) (200, 91.7)};
\addplot[color=color2,mark=triangle,densely dotted,mark options={solid}] coordinates {(0, 50.9) (10, 47.5) (25, 59.1) (50, 59.6) (75, 63.4) (100, 68.8) (125, 65.2) (150, 71.1) (175, 66.8) (200, 67.8)};
\addplot[color=color3,mark=*] coordinates {(0, 39.5) (10, 45.8) (25, 50.3) (50, 52.9) (75, 54.2) (100, 54.9) (125, 58.9) (150, 60.8) (175, 61.5) (200, 64.4)};
\addplot[
    only marks,
    visualization depends on=\thisrow{alignment} \as \alignment,
    nodes near coords,
    point meta=explicit symbolic,
    every node near coord/.style={anchor=\alignment, font=\tiny}
    ] table [
     meta index=2
     ] { x       y       label       alignment
     0 61.1 61.1 -55
     0 50.9 50.9 -90
     0 39.5 39.5 90
     200 91.7 91.7 90
     200 67.8 67.8 -90
     200 64.4 64.4 100
     };

\legend{SPTE, MPTE, EPC}

\end{axis}
\end{tikzpicture}

\vspace{-0.1in}
\caption{Learning curves of the best models on each RNPC task with an increasing number of finetuning examples.}
\label{figure:inoculation_results}
 \vspace{-0.1in}
\end{figure}

On SPTE, the accuracy rapidly climbs from 61.1 to 75.8 with only 10 examples, and saturates around 92 with 100 examples, approaching human performance (94.1). The learning curve on MPTE has more fluctuations, with a peak at 71.1 (150 examples) and a final score of 67.8. On EPC, starting around chance (39.5), the accuracy progressively increases up to 64.4 with 200 examples. These results indicate that the target knowledge is learnable with appropriate training data. Furthermore, SPTE may be the easiest task, since it only requires local knowledge about the meaning of the modifiers and the noun. By contrast, MPTE and EPC involve world knowledge (e.g., basketball players are generally tall among the population), as well as global reasoning between components in a sentence (e.g., the relationship between the event and the modifiers), which may explain the remaining large gap between model and human performance ($>90$).

\vspace{-0.08in}
\section{What can LMs learn from RNPC?}
\label{section:Qc}
\vspace{-0.08in}

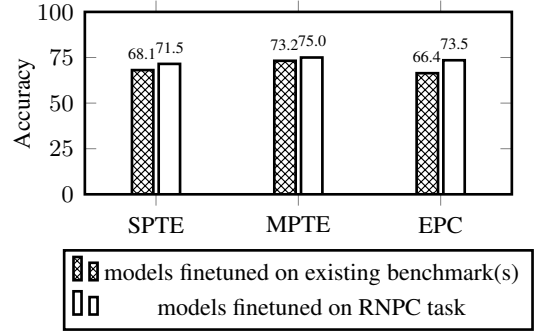
\begin{figure}[t!]
\centering
\begin{tikzpicture}
\begin{axis}[
    height=4cm,
    width=0.45\textwidth,
    symbolic x coords={SPTE, MPTE, EPC},
    xtick=data,
    xtick align=center,
    x tick label style={font=\footnotesize},
    enlarge x limits=0.25,
	ylabel={Accuracy},
	y label style={font=\small, yshift=-0.25cm},
	y tick label style={font=\footnotesize},
	ymin=0,
	ymax=100,
	ytick distance=25,
	legend style={legend columns=1, font=\small},	
	ybar,
	bar width=8pt,
	legend style={at={(0.5,-0.3)},anchor=north,legend columns=1, font=\footnotesize},
	]
\addplot [pattern=crosshatch]
	coordinates {(SPTE, 68.1) (MPTE, 73.2) (EPC, 66.4)};
\addplot [color=black, fill=white]
	coordinates {(SPTE, 71.5) (MPTE, 75.0) (EPC, 73.5)};
\addplot[
    only marks,
    visualization depends on=\thisrow{alignment} \as \alignment,
    nodes near coords,
    point meta=explicit symbolic,
    every node near coord/.style={anchor=\alignment, font=\tiny}
    ] table [
     meta index=2
     ] { x       y       label       alignment
     SPTE 68.1 68.1 -50
     SPTE 71.5 71.5 230
     MPTE 73.2 73.2 -50
     MPTE 75.0 75.0 230
     EPC 66.4 66.4 -50
     EPC 73.5 73.5 230
     };
 \legend{models finetuned on existing benchmark(s), models finetuned on RNPC task}
 
\end{axis}

\end{tikzpicture}
\vspace{-0.05in}
\caption{Probing accuracy for the ``modifier semantic category'' feature, before (left) and after (right) finetuning on each RNPC task.}\label{figure:mod_class_probing_results}
 \vspace{-0.1in}
\end{figure}

Given that the target knowledge is learnable, we now address question (c): What linguistic features have the models learned from RNPC? We probe for two features extensively studied in the relevant literature (cf. \cref{sec:related_work}), using different techniques.

\minisection{Modifier semantic category.}
We first investigate if models have learned the semantic category of modifiers using the ``edge probing technique'' \cite{tenney2018what}. Namely, each modifier is categorized as intersective, subsective, or privative \cite{mccrae-etal-2014-modelling}. The entailment pattern of individual modifiers is an important factor in determining the meaning of the entire NP.

Given a finetuned model, we take the contextualized representation of each modifier in the last hidden layer. Then, we attach a linear head on top of the token representation as an ``auxiliary classifier''. We choose linear classifiers because more expressive ones like Multi-Layer Perceptron are more likely to capture the target feature themselves \cite{hewitt-liang-2019-designing}. The token representations are then frozen, while the linear head is trained to predict the semantic category of the modifiers.\footnote{See Appendix~\ref{appendix:edge_probing} for an illustration of the method.} 

We probe the models finetuned on RNPC from Section~\ref{section:Qb}, as well as the models finetuned on existing benchmarks for comparison. The results are shown in Figure~\ref{figure:mod_class_probing_results}. For all tasks, the probing accuracy is higher for models finetuned on RNPC than on existing benchmarks. The increase is small for SPTE (3.4) and MPTE (2.8), but more obvious for EPC (7.1). This is somewhat counter-intuitive since modifier category is defined in terms of entailment patterns, but models learn it better from EPC than from TE tasks. Nonetheless, the overall trend shows that models can learn the semantic category of modifiers to some extent after being finetuned on our datasets. Since the absolute increase is limited, we plan to explore ways to quantify the actual amount of learned knowledge in future work. 

\begin{figure}[t]
    \centering
    \includegraphics[width=\linewidth, height=6.2cm]{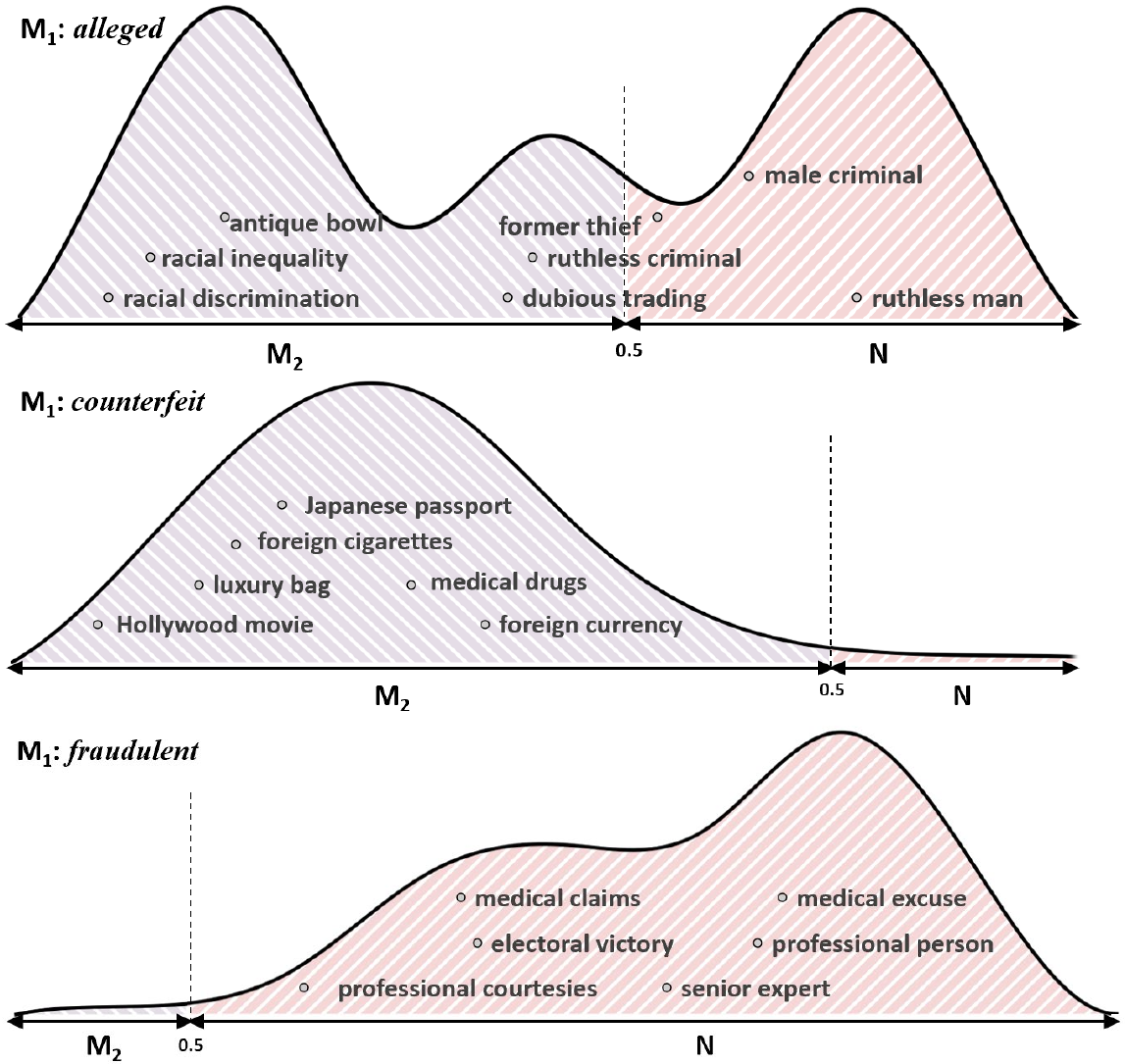}
    \caption{A case study of modifier scope. Each sub-figure shows the frequency distribution of the attention ratio $r$ ($0 < r < 1$) for an $\mathbf{M_1}$, divided into two sides at $0.5$. The $\mathbf{M_2}$ side contains NPs where $\mathbf{M_1}$ attends more to $\mathbf{M_2}$ than to $\mathbf{N}$; vice versa for the $\mathbf{N}$ side. 
    }
    \vspace{-0.1in}
    \label{figure:attention}
\end{figure}

\minisection{Modifier scope.} We also probe for the scope of the first modifier ($\mathbf{M_1}$) in recursive NPs ($\mathbf{Det\ M_1\ M_2\ N}$). Specifically, we focus on privative $\mathbf{M_1}$'s, since they can have different scopes when interacting with different $\mathbf{M_2}$'s and $\mathbf{N}$'s. For instance, 
in the NP \textit{a former American diplomat}, {\it former} negates {\it diplomat} ($\mathbf{N}$), but the person is still American; while in \textit{a former beginner drummer}, it negates \textit{beginner} ($\mathbf{M_2}$), but the person may still be a drummer.\footnote{Admittedly, there can be alternative interpretations: say, one can also imagine that \textit{a former beginner drummer} describes a person who is no longer a drummer at all. However, in that case, it is enough to say \textit{a former drummer} instead, considering the Gricean maxim of quantity. Therefore, here we still focus on the first interpretation, which is more straightforward.} This difference cannot be captured by the semantic category of \textit{former}.

As a proxy for the scope of $\mathbf{M_1}$, we use attention visualization, a widely adopted technique to study token correlations~\cite{vig-2019-multiscale}.\footnote{There have been recent debates on the faithfulness of this method \cite{jain-wallace-2019-attention, wiegreffe-pinter-2019-attention}. Therefore, we do not use attention weights to make claims about \textbf{how} our models work, but only \textbf{what} they capture, with attention weights.} We choose BERT-base finetuned on 200 MPTE examples from Section~\ref{section:Qb} as the model to be probed for a case study. 

Let us denote any token in a given NP as $\mathbf{x}$. We define $\mathbf{A_{x}}$, the average of the weights of all attention heads from $\mathbf{M_1}$ to $\mathbf{x}$ in the final layer, representing how much $\mathbf{M_1}$ attends to token $\mathbf{x}$. We then calculate the ratio $r = \mathbf{A_{N} / (A_{N} + A_{M_2})}$ ($0 < r < 1$). If $r < 0.5$, then $\mathbf{M_1}$ attends more to $\mathbf{M_2}$; else, $\mathbf{M_1}$ attends more to $\mathbf{N}$. For each privative modifier, we take all NPs containing it in the $M_1$ position in our dataset and plot the distribution of $r$. Figure~\ref{figure:attention} shows three examples (\textit{alleged}, \textit{counterfeit}, or \textit{fraudulent}) representing different patterns.

As shown in the first sub-figure, \textit{alleged} attends more to either $\mathbf{M_2}$ and $\mathbf{N}$ depending on the NP. For example, it attends more to $\mathbf{M_2}$ in \textit{an alleged antique bowl} (0.454), since the NP describes a bowl that may not be antique. Inversely, \textit{an alleged male criminal} is on the $\mathbf{N}$ side (0.517), since 
they are most likely male but may not be a criminal. 

The second sub-figure indicates that \textit{counterfeit} mainly attends to $\mathbf{M_2}$. For instance, \textit{a counterfeit Hollywood movie} (0.382) is still a movie, but is probably not made in Hollywood. This is similar to the cases of \textit{luxury bag}, \textit{medical drugs}, \textit{foreign cigarettes}, etc. On the contrary, \textit{fraudulent} mainly attends to $\mathbf{N}$, as shown in the third sub-figure. The \textit{fraudulent medical claims} (0.559) are not valid claims but still on  medical grounds. The same holds for \textit{electoral victory}, \textit{medical excuse}, etc.

Additionally, we notice that there are some boundary cases close to the $\mathbf{r=0.5}$ division line, like \textit{ruthless criminal} and \textit{former thief} in the \textit{alleged} sub-figure. A plausible explanation is that $\mathbf{M_1}$ is questioning both $\mathbf{M_2}$ and $\mathbf{N}$ in these cases (e.g., \textit{an alleged ruthless criminal} is not necessarily ruthless or a criminal). Overall, the above results indicate that models finetuned on our tasks can capture modifier scope in recursive NPs.



\vspace{-0.05in}
\section{Is RNPC useful for downstream tasks?}
\label{section:Qd}
\vspace{-0.05in}

We finally address question (d): How can such knowledge benefit downstream tasks? We choose the task of Harm Detection \cite{banko-etal-2020-unified} for extrinsic evaluation. Concretely, we consider the scenario where a user interacts with a task-oriented agent like Siri or Alexa, and the agent needs to determine whether the involved activity in the user query is potentially harmful. The definition of ``harm'' can be user-dependent. Here, we consider an activity to be harmful if it may cause pain, physical injury, or be illegal for \textbf{minors}. We choose this task because many false positives come from recursive NPs. For example, \textit{how to make a homemade bomb} is obviously \texttt{harmful} while \textit{how to make a homemade bath bomb} is \texttt{harmless}.

We collect a small test set from wikiHow, a website of how-to articles. Each article title is considered a query (e.g., \textit{how to make a cake}). Then, we compile a list of 74 keywords about harmful entities (e.g., \textit{bomb}, \textit{fire}, \textit{drugs}), only 12 of which occur in RNPC. We then select wikiHow queries containing at least an NP with one of the 74 keywords as the head noun, and sample a small subset for manual annotation. Each query is labeled as \texttt{harmful} or \texttt{harmless}, depending on whether it involves a harmful activity as defined above. After data cleaning and re-balancing, we obtain 170 queries, with a 1:1 positive/negative ratio.

We design two zero-shot harm classifiers using models finetuned on our entire SPTE and EPC dataset. They share a few pre-processing steps: first, all NPs are extracted from the input query; then, NPs containing a keyword from our list in the head noun position are retained. 
For each retained NP (e.g., \textit{a water gun}), we check if it is indeed a harmful entity using either the SPTE or the EPC model. The input to the SPTE model is a premise of the form ``This is \{NP\}'' (e.g., \textit{This is a water gun}) and a hypothesis of the form ``This is (a/an) \{N\}'' (e.g., \textit{This is a gun}). If the output label is \texttt{entailment}, we classify the query as \texttt{harmful}, otherwise \texttt{harmless}. Likewise, using the EPC model, we form two events given the retained NP: ``(A/An) \{N\} is harmful'' and ``\{NP\} is harmful''. If the second event is predicted as more or equally plausible compared to the first, the query is considered \texttt{harmful}.

\begin{table}[!t]
\centering
\scalebox{0.8}{%
\begin{tabular}{lcccc}
    \hline \textbf{Model} & \textbf{Acc.}  & \textbf{P}  & \textbf{R} & $\mathbf{F_1}$ \\
    \hline Always \texttt{harmful} & 50.0 & 50.0 & \textbf{100.0} & 66.7 \\
    GPT3-ada & 49.4 & 49.7 & 98.8 & 66.1 \\
    GPT3-curie & 59.4 & 60.5 & 54.1 & 57.1 \\
    GPT3-davinci & 51.3 & 50.6 & \textbf{100.0} & 67.2 \\
    RoBERTa-large (SPTE) (\textbf{ours}) & 58.2 & 54.5 & \textbf{100.0} & 70.5 \\
    RoBERTa-large (EPC) (\textbf{ours}) & \textbf{72.9} & \textbf{66.4} & 92.9 & \textbf{77.5}\\
    
    \hline
\end{tabular}
}
 \vspace{-0.05in}
\caption{Zero-shot performance of models trained on RNPC on the Harm Detection task. Baselines include a model that always predicts \texttt{harmful} and GPT3.}\label{table:transfer_results}
\vspace{-0.15in}
\end{table}

We compare our two classifiers to a simple baseline that always predicts \texttt{harmful} as well as to three GPT3 models.\footnote{Used in a zero-shot setting; see Appendix~\ref{appendix:implementation_transfer_model} for details.} Both classifiers meaningfully exceed the simple baseline, and the EPC-based classifier outperforms all the other methods by 10+ in terms of accuracy and $F_1$. This shows that the understanding of recursive NPs is beneficial for downstream tasks without any training data.
To understand why EPC is more suitable than SPTE for this task, we further examine the errors they make. One major error type concerns polysemous keywords such as \textit{shot}. For instance, the SPTE model mistakenly predicts \textit{how to have a good basketball shot} to be \texttt{harmful} because \textit{a good basketball shot} is still a \textit{shot} (\textit{shot} can mean both ``shooting a gun'' and ``shooting a ball''). There are also some queries out of the scope of the EPC model, e.g., \textit{how to make a sake bomb}. Since \textit{sake bomb} is a cocktail, the gold label is \texttt{harmful} as our target users are minors. The EPC model correctly predicts that \textit{a sake bomb} is less harmful than \textit{a bomb}, but fails to capture that it may still be harmful (for minors). 

\vspace{-0.05in}
\section{Conclusion}
\vspace{-0.05in}

We introduce RNPC, a challenge set targeting the understanding of recursive NPs, a fundamental aspect of human common sense. Pretrained LMs with SOTA performance on Natural Language Understanding benchmarks have poor mastery of this knowledge, but can still learn it when exposed to small amounts of data from RNPC. 
Using different probing techniques, we show that models can learn relevant linguistic features, including modifier category and scope, from RNPC. They also achieve strong zero-shot performance on an extrinsic Harm Detection task, indicating the transferability of this knowledge. For future work, we hope to investigate other linguistic phenomena as a step towards comprehensively characterizing LMs' limitations and capabilities in language understanding.

\section*{Acknowledgments}
This research is based upon work supported in part by the DARPA KAIROS Program (contract FA8750-19-2-1004), the DARPA LwLL Program (contract FA8750-19-2-0201), and the IARPA BETTER Program (contract 2019-19051600004). Approved for Public Release, Distribution Unlimited. The views and conclusions contained herein are those of the authors and should not be interpreted as necessarily representing the official policies, either expressed or implied, of DARPA, IARPA, or the U.S. Government.

Special thanks go to our annotators, students in CIS 421/521 and MCIT 521 at the University of Pennsylvania. We also thank Artemis Panagopoulou for providing the extrinsic evaluation data. Meanwhile, we appreciate the support from OpenAI on finetuning GPT-3. Finally, we thank Haochen Zhang, Pengyuan Lu, Daniel Deutsch, Daphne Ippolito, Lara Martin, Young-Min Cho, Yi Zhang, Helen Jin, Siyi Liu, Eleni Miltsakaki, Jordan Kodner, Mingming Liu, Peng Zhou, Christopher Cieri, James J. Fiumara, Ellie Pavlick, Charles Yang, Yejin Choi, Alexander Koller, Chris Potts, and Mitch Marcus for their valuable feedback.


\bibliography{custom}
\bibliographystyle{acl_natbib}

\clearpage
\appendix

\section{Dataset Construction Details}
\label{appendix:data_construction}
\subsection{RNPC Statistics}

\minisection{NPs.} RNPC has 1,299 NPs. For an NP in the form of $\mathbf{Det\ M_1\ M_2\ N}$, the two modifiers $M_1$ and $M_2$ can each belong to one of three possible semantic categories (intersective, subsective, or privative), resulting in nine possible combinations. We plot the distribution of NPs with different combinations in RNPC in Table~\ref{table:combo_dist}. Note that the distribution is not balanced because certain categories (e.g., NPs containing privative modifiers) yield many more minority class examples for our three tasks (e.g., \texttt{non-entailment} in SPTE). Thus, considering the final class balance in RNPC tasks, we include more NPs of certain categories. 

\minisection{Training and test sets for finetuning.}
In the experiment where we finetune models on RNPC,  described in Section~\ref{section:Qb}, we split again the dataset for each task into a training set and a new test set, ensuring no overlap of modifiers occurring in the same position. The training set contains 200 examples, which are gradually provided to the model. The test set contains the remaining examples. Table~\ref{table:ino_split} shows the number of examples for each task.

\subsection{Crowdsourcing Details}

In the construction of RNPC, we hire college students as crowdworkers for instance creation and label verification. Specifically, they are undergraduate and graduate students in an Artificial Intelligence class (CIS 421/521 and MCIT 521 at the University of Pennsylvania), with good English proficiency. Both tasks are given as optional extra credit assignments in the class. Participation is solely voluntary. Before participation, students can preview the tasks, and are given a clear description of how the data will be used at the beginning of the instructions.

During instance creation, we provide detailed instructions on how to write high-quality examples for each task, which can be found in the Supplementary Materials. Annotations are collected via Google Forms. With 100 valid instances (equivalent to 2.5-4.75 hours of work, depending on their proficiency), students can earn 1\% in extra credit of the overall course grade.

\begin{table}[t]
\begin{center}
\scalebox{0.9}{
\begin{tabular}{l|rrr} 
\hline $\mathbf{M_1}$ / $\mathbf{M_2}$  & \bf Int. & \bf Sub. & \bf Pri. \\ 
\hline \bf Int. & 13 & 37 & 74 \\
\bf Sub. & 138 & 109 & 162 \\
\bf Pri. & 99 & 420 & 250 \\

\hline
\end{tabular}
}
\end{center}
 \vspace{-0.05in}
\caption{Number of NPs in RNPC with different combinations of modifier category in the $M_1$ and $M_2$ position. Possible categories include intersective, subsective, and privative.}\label{table:combo_dist}
\end{table}

\begin{table}[t!]
\begin{center}
\scalebox{0.9}{
\begin{tabular}{l|rr} 
\hline \bf Task & \bf Train & \bf Test \\ 
\hline \bf SPTE & 200 & 963 \\
\bf MPTE & 200 & 863 \\
\bf EPC & 200 & 1,279 \\
\hline
\end{tabular}
}
\end{center}
 \vspace{-0.05in}
\caption{Number of examples in the training and testing split for each RNPC task in the finetuning experiment.}\label{table:ino_split}
 \vspace{-0.15in}
\end{table}

During label verification, we host our questions on Amazon Mechanical Turk. We design a HIT type for each RNPC task, which is also included in the Supplementary Materials. With 600 correctly answered questions (equivalent to 3.5-4 hours of work), students can earn 1\% in extra credit of the overall course grade. We calculate the inter-annotator agreement using Krippendorff's alpha.\footnote{\url{https://pypi.org/project/krippendorff}} The agreement is 0.843 for SPTE, 0.575 for MPTE, and 0.933 for EPC.

\subsection{Debiasing and Anonymization} The collected data does not contain any information that names or uniquely identifies individual people or offensive content. We ensure this by 1) manually reviewing the set of extracted NPs from corpora, and filtering out any NP that contains any sensitive/offensive information, 2) not requesting any personal information during human annotation, and 3) manually reviewing each RNPC example written by the human participants.

\section{Existing Benchmarks for Finetuning}
\label{appendix:benchmark_details}

We use the following benchmark datasets 
for finetuning. Each of them has the same format as one of our RNPC tasks. Table~\ref{table:existing_benchmark_stats} shows the number of examples in each dataset.

\begin{table}[!t]
\begin{center}
\scalebox{0.9}{
\begin{tabular}{l|rrr} 
\hline \bf Dataset & \bf Train & \bf Dev & \bf Test \\ 
\hline MNLI & 392,702 & 20,000 & 20,000 \\
SNLI & 550,152 & 10,000 & 10,000 \\
MPE & 8,000 & 1,000 & 1,000 \\
ADEPT & 12,892 &  1,611 & 1,612 \\

\hline
\end{tabular}
}
\end{center}
 \vspace{-0.05in}
\caption{Number of examples in existing datasets of the same format used for finetuning.}\label{table:existing_benchmark_stats}
\end{table}

\minisection{MNLI.} The Multi-Genre Natural Language Inference corpus \cite{williams2018broad} is a dataset of 433k textual entailment examples, labeled as entailment, contradiction, or neutral. It covers a range of genres of spoken and written text. The language in the dataset is English. The corpus is released under the OANC’s license, the Creative Commons Share-Alike 3.0 Unported License, and the Creative Commons Attribution 3.0 Unported Licenses, depending on the portion.

\minisection{SNLI.} The Stanford Natural Language Inference corpus \cite{bowman-etal-2015-large} is a crowdsourced dataset of textual entailment examples, labeled as entailment, contradiction, or neutral. The sentences are written by humans doing a novel grounded task based on image captioning. The language in the dataset is English. The dataset is released under the Creative Commons Attribution-ShareAlike 4.0 International License.

\minisection{MPE.} \citet{lai-etal-2017-natural} introduce a Multiple Premise Entailment Task dataset. This is a novel textual entailment task that requires inference over multiple premise sentences. Each example consists of four premise sentences (captions from a FLICKR30K image), one hypothesis sentence (a simplified FLICKR30K caption), and one label (entailment, neutral, or contradiction) that indicates the relationship between the set of four premises and the hypothesis. The language in the dataset is English. The license of the dataset is unspecified.  

\minisection{ADEPT.} \citet{emami-etal-2021-adept} introduce a dataset of the Adjective-Dependent Plausibility Task (ADEPT). Each example contains a base sentence, and a slightly modified sentence obtained by adding an adjective to a noun in the base sentence. The dataset is created to support explorations into how certain classes of adjectives might influence the plausibility of events depicted in natural language sentences. The textual data come from Wikipedia, the Common Crawl, and ConceptNet. The language of the dataset is English. ADEPT is released under the CC BY-SA 3.0 license. It is intended to be used only for research, exploratory evaluation, and auditing, which our use is consistent with.

\section{Probing Pretrained LMs}
\label{appendix:direct_probe}

\subsection{Motivation} 
When addressing question (a), we finetune pretrained LMs on existing benchmarks of the same format as each RNPC task, assuming that the finetuning process allows models to do textual inference in the required format. However, it is possible that this assumption does not hold, because LMs can overfit the finetuning data beyond just learning the format. Then even if the target knowledge is present in pretrained LMs, \textit{catastrophic forgetting} \cite{kemker2018measuring} can happen during finetuning.

\subsection{Task Conversion} 
We complement Section~\ref{section:Qa} with another experiment, where we directly probe pretrained LMs using a prompting method inspired by the line of work on LMs as knowledge bases \cite{petroni2019language}. Specifically, we convert each RNPC task to a likelihood comparison task:

\minisection{SPTE.} Given the original formulation which has a premise and a hypothesis, we define $L_{entail}$ as the \textbf{conditional likelihood} that the hypothesis is necessarily true given the premise, assigned by an LM. Contrarily, $L_{non-entail}$ stands for the conditional likelihood that the hypothesis is NOT necessarily true given the premise.\footnote{For example, if the original SPTE example has the premise \textit{This is the second green ball} and the hypothesis \textit{This is the second ball}, then $L_{entail}$ equals to $L$(\textit{This is necessarily the second ball} | \textit{This is the second green ball}), and $L_{non-entail}$ equals to $L$(\textit{This \textbf{isn't} necessarily the second ball} | \textit{This is the second green ball}).} If $L_{entail} > L_{non-entail}$, the model is considered to predict \texttt{entailment}, and vice versa.

\minisection{MPTE.} The conversion method is the same as that for SPTE, except that in the conditional likelihood computation, we now consider the concatenation of two premises as the given condition.

\minisection{EPC.} Given the original formulation with two events, Event 1 and Event 2, we define $L_{1}$ and $L_{2}$ as the \textbf{(unconditional) likelihood} of Event 1 and Event 2 assigned by an LM, respectively. We then choose a threshold $\theta$,\footnote{In the range [0.1, 0.5, 1, 2, 3, 5], 0.5 is the empirical optimal.} and compare it to the absolute difference between $L_{1}$ and $L_{2}$. If the difference is smaller than $\theta$, we consider the model prediction as \texttt{equally likely}. Otherwise, the model prediction is \texttt{more likely} if $L_{2}$ is higher, and \texttt{less likely} if $L_{1}$ is higher.

For Causal LMs (e.g., GPT), the likelihood is computed with standard left-to-right language modeling scores. For Masked LMs (e.g., BERT, RoBERTa, BART), the likelihood is computed with pseudo-log-likelihood scores \cite{salazar2020masked}. 

\subsection{Sanity Check}
Before evaluating LMs on the converted RNPC, we perform a sanity check to see if our formalization makes sense to LMs, i.e., whether they understand the meaning of \textit{necessarily} and \textit{not necessarily}.

We write 50 sentence pairs for likelihood comparison, all consisting of simple commonsense knowledge. For example, comparing \textit{A human being is necessarily female} and \textit{A human being isn't necessarily female}, the second sentence should be more likely; while for \textit{Humans are necessarily mortal} and \textit{Humans aren't necessarily mortal}, the first sentence should be more likely. Such comparisons do not require any knowledge about recursive NPs, and involve only common entities and facts. If models understand \textit{necessarily} and \textit{not necessarily} correctly, they should find the task easy.

\begin{table}[!t]
\centering
\scalebox{0.9}{%
\begin{tabular}{lccc}
    \hline \textbf{Model} & \textbf{SPTE}  & \textbf{MPTE}  & \textbf{EPC}  \\
    \hline gpt2-base & 59.4 & 52.8 & 33.4 \\
    gpt2-medium & \textbf{62.6} & 53.6 & 33.6 \\
    gpt2-large & 61.4 & 56.1 & 32.4 \\
    gpt2-xl & 61.7 & \textbf{56.9} & 31.7 \\
    gpt3-ada & 55.2 & 55.2 & 33.2\\
    \hline
\end{tabular}
}
 \vspace{-0.05in}
\caption{Accuracy of SOTA pretrained models directly evaluated on RNPC tasks.}\label{table:direct_probing_results}
 \vspace{-0.1in}
\end{table}

To our surprise, almost all Masked LMs we test (BERT-base/large, RoBERTa-base/large) fail the sanity check, mostly performing around chance (50 accuracy). However, most Causal LMs (GPT-2-base/medium/large/xl, GPT-3-ada) reasonably perform above chance, with accuracy scores ranging from 70 to 80. We suspect that pseudo-log-likelihood scores are not entirely suitable for our purposes; also, the task is harder than expected due to reporting bias, as the tested knowledge (e.g., \textit{not all humans are female}) is potentially too obvious to be explicitly stated in the pretraining data. 

\subsection{Results}
We evaluate LMs that pass the sanity check on the converted RNPC, and report their performance in Table~\ref{table:direct_probing_results}. Despite the decent performance on the sanity check examples (70-80), the accuracy on RNPC is remarkably lower. Compared to our original results of probing the finetuned models, the optimal performance on SPTE and MPTE slightly improves, while accuracy on EPC decreases. However, the same patterns hold: most models perform around or slightly above chance, with a large difference from human performance. These findings further strengthen our answer to question (a), i.e. LMs do not inherently have the knowledge to interpret recursive NPs.

\section{Full Results}
In Section~\ref{section:Qa}, we evaluate SOTA LMs on RNPC tasks. In addition to accuracy, we also report precision, recall, and F-1 score here. Tables~\ref{table:full_results_SPTE},~\ref{table:full_results_MPTE} and ~\ref{table:full_results_EPC} show the full results for each task, respectively.

\begin{table}[!t]
\centering
\scalebox{0.85}{%
\begin{tabular}{lcccc}
    \hline \textbf{Model} & \textbf{Acc.}  & \textbf{P}  & \textbf{R} & $\mathbf{F_1}$ \\
    \hline  BERT-base (SNLI) & 49.8 & 49.9 & 77.0 & 60.5 \\
    BERT-base (MNLI) & 51.3 & 50.7 & 97.8 & 66.8 \\
    RoBERTa-large (MNLI) & \textbf{61.1} & 56.3 & 99.1 & 71.9 \\
    BART-large (MNLI) & 59.3 & 55.1 & 97.9 & 70.7 \\
    \hline
\end{tabular}
}
 \vspace{-0.05in}
\caption{Full results of SOTA models evaluated on SPTE. The finetuning dataset is in brackets.}\label{table:full_results_SPTE}
 \vspace{-0.05in}
\end{table}

\begin{table}[!t]
\centering
\scalebox{0.85}{%
\begin{tabular}{lcccc}
    \hline \textbf{Model} & \textbf{Acc.}  & \textbf{P}  & \textbf{R} & $\mathbf{F_1}$ \\
    \hline  BERT-base & 47.2 & 48.0 & 44.0 & 45.9 \\
    BERT-large & 41.5 & 34.2 & 16.3 & 22.1 \\
    RoBERTa-base & 51.1 & 51.0 & 100.0 & 67.5 \\
    RoBERTa-large  & 50.9 & 50.9 & 100.0 & 67.5 \\
    GPT3-ada & 52.0 & 51.5 & 97.0 & 67.3 \\
    GPT3-curie & \textbf{54.1} & 52.6 & 97.4 & \textbf{68.4} \\
    \hline
\end{tabular}
}
 \vspace{-0.05in}
\caption{Full results of SOTA models evaluated on MPTE. The finetuning dataset is MPE for all models.}\label{table:full_results_MPTE}
\vspace{-0.05in}
\end{table}

\begin{table}[!t]
\centering
\scalebox{0.85}{%
\begin{tabular}{lcccc}
    \hline \textbf{Model} & \textbf{Acc.}  & \textbf{P}  & \textbf{R} & $\mathbf{F_1}$ \\
    \hline BERT-base & 31.6 & 29.2 & 31.6 & 22.4 \\
    BERT-large & 32.2 & 27.7 & 32.2 & 23.7 \\
    RoBERTa-base & 31.0 & 46.8 & 31.0 & 22.3 \\
    RoBERTa-large & \textbf{39.5} & 54.1 & 39.5 & \textbf{32.7} \\
    GPT3-ada & 35.2 & 40.2 & 35.2 & 28.3 \\
    GPT3-curie & 38.7 & 69.9 & 38.7 & 32.8 \\
    \hline
\end{tabular}
}
 \vspace{-0.05in}
\caption{Full results of SOTA models evaluated on EPC. The finetuning dataset is ADEPT for all models.}\label{table:full_results_EPC}
 \vspace{-0.05in}
\end{table}

\section{Implementation Details}
\subsection{Models Finetuned on Existing Benchmarks}
\label{appendix:implementation_eval_model}
In Section~\ref{section:Qa}, we evaluate SOTA LMs finetuned on existing benchmarks of the same format on RNPC. We use four different pretrained models, BERT \cite{devlin-etal-2019-bert}, RoBERTa \cite{liu2019roberta}, BART \cite{lewis-etal-2020-bart}, and GPT3 \cite{NEURIPS2020_1457c0d6}, in different sizes. The first three are implemented with HuggingFace Transformers\footnote{\url{https://github.com/huggingface/transformers}}, and the last is from OpenAI's standard API\footnote{\url{https://beta.openai.com/docs/api-reference}}.

The pretrained model checkpoints we use include: \texttt{bert-base-uncased} (110M parameters), \texttt{bert-large-uncased} (336M parameters), \texttt{roberta-base} (125M parameters), \texttt{roberta-large} (335M parameters),  \texttt{facebook/bart-large} (406M parameters), \texttt{GPT3-ada} (350M parameters), and \texttt{GPT3-curie} (6.7B parameters).\footnote{All models above are available at \url{https://huggingface.co/transformers/v4.8.2/pretrained_models.html} or \url{https://beta.openai.com}} Their licenses include Apache License 2.0 (BERT and BART), GNU General Public License v2.0 (RoBERTa), and MIT license (GPT3).

Due to the size of MNLI and SNLI, we use existing checkpoints available on the Huggingface Transformers model hub. For all other datasets, we finetune the pretrained models using the \texttt{SequenceClassification} pipeline on Huggingface, or the standard prompt completion finetuning API on OpenAI.\footnote{\url{https://beta.openai.com/docs/api-reference/fine-tunes}}  The finetuning scripts are adapted from the \texttt{text-classification} example in the HuggingFace Transformers repository.\footnote{\url{https://github.com/huggingface/transformers/tree/master/examples/legacy}} We performed hyperparameter search in the following range:\\
\indent - batch size: [4, 8, 16, 32]\\
\indent - learning rate: [1e-5, 1e-6]\\
\indent - number of epochs: [2, 3, 5]\\
\indent - max sequence length: [64, 128]

The optimal hyperparameter values and finetuned models are available on the HuggingFace model hub.

We run our finetuning experiments on an NVIDIA GeForce RTX 2080 Ti GPU, with half-precision floating point format (FP16). The finetuning takes 2 to 5 hours depending on the task.

\subsection{Models Finetuned on RNPC} 
\label{appendix:implementation_ino_model}

In Section~\ref{section:Qb}, we address the question of whether LMs can learn the meaning of recursive NPs. We finetune each model from Section~\ref{appendix:implementation_eval_model} on an increasing number of examples of each RNPC task. The model architectures, the pipelines used, the range of hyperparameter search, and the computing resources used are all the same as in the previous subsection. After being finetuned on 200 examples, the best performing models are RoBERTa-large (MNLI) for SPTE, RoBERTa-base (MPE) for MPTE, and RoBERTA-large (ADEPT) for EPC. The optimal hyperparameter values and finetuned models on the full 200 examples of each RNPC task are available on the HuggingFace model hub.

\subsection{The ``Edge Probing'' Method}
\label{appendix:edge_probing}

\begin{figure}[!t]
\includegraphics[width=\columnwidth]{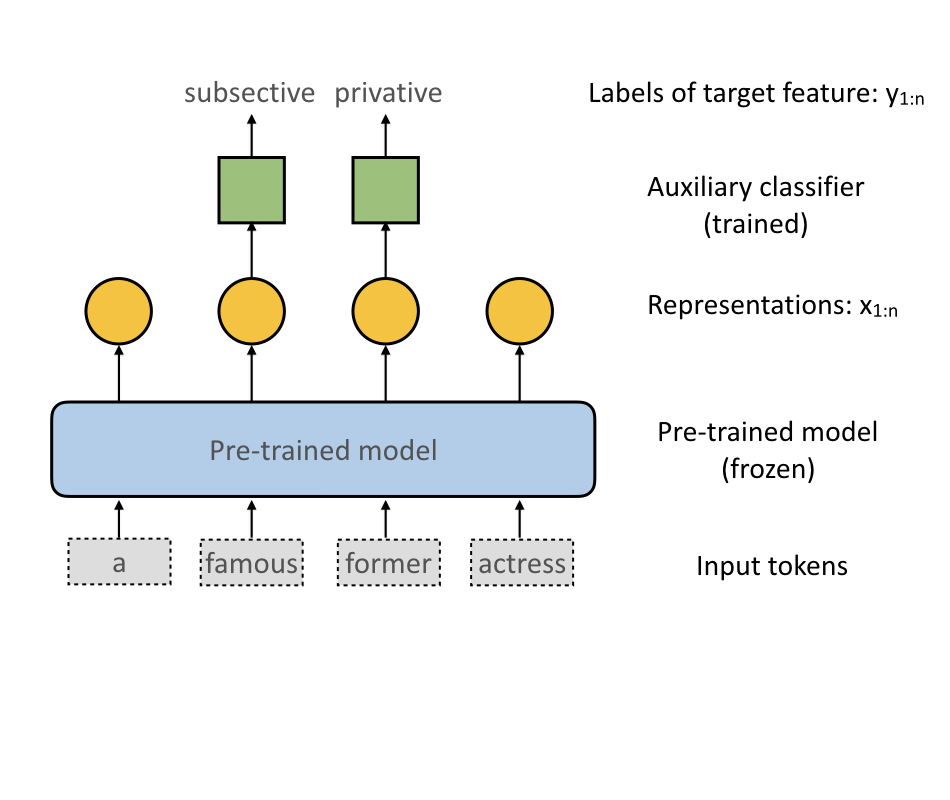}
\vspace{-0.6in}
\caption{An illustration of the Edge Probing method. Figure adapted from \citet{tenney2018what}.}\label{figure:edge_probing}
\vspace{-0.1in}
\end{figure}

In Section~\ref{section:Qc}, we adopt the Edge Probing technique from \citet{tenney2018what} to investigate if the modifier category feature can be learned from our tasks. 

To reintroduce the general idea of this method, consider the following setup: we have data $D=\{(x_1, y_1), (x_2, y_2), ..., (x_n, y_n)\}$, where $(x_1, x_2,..., x_n)$ are the model representations to be probed and $(y_1, y_2,..., y_n)$ are the labels of a linguistic feature we are interested in probing for. The goal is to see if $(x_1, x_2,..., x_n)$ encodes $(y_1, y_2,..., y_n)$. 

In our case, given an NP of the form $\mathbf{Det\ M_1\ M_2\ N}$, $(x_1, x_2,..., x_n)$ are the token representations of the best-performing models after being finetuned on each RNPC task, as mentioned in Section~\ref{appendix:implementation_ino_model}, and $(y_1, y_2,..., y_n)$ are the semantic categories of $\mathbf{M_1}$ and $\mathbf{M_2}$.

We freeze the representations $(x_1, x_2,..., x_n)$, and attach a simple auxiliary classifier (or probing classifier) on top of them. The auxiliary classifier is trained to predict the semantic category $y_i$ corresponding to every modifier token representation $x_i$. In our case, we choose a linear classifier as the probing classifier. The assumption is that the expressivity of the auxiliary classifier should be limited, or otherwise it might learn the target feature itself. Based on this assumption, the accuracy of predicting $y$ indicates the extent to which the model representations encode the target feature. The method is illustrated in Figure~\ref{figure:edge_probing}. 

Specifically, the linear classifier is an \texttt{SGDClassifier} implemented with scikit-learn.\footnote{\url{https://scikit-learn.org}} We perform hyperparameter search in the following range:\\
\indent - loss: [hinge, log, modified\_huber, squared\_hinge, perceptron]\\
\indent - warm\_start: [True, False]\\
\indent - penalty: [l2, l1, elasticnet]\\
\indent - alpha: [0.0001, 0.001, 0.01, 0.1]\\
\indent - n\_jobs: [-1]\\
\indent - max\_iter: [500, 1000, 2000]\\
\indent - random\_state: [0]\\
\indent - eta0: [0.1, 0.01, 0.001, 0.0001]\\
\indent - learning\_rate: [optimal, invscaling, adaptive]

The optimal hyperparameter values for the best-performing models of each RNPC task from Section~\ref{appendix:implementation_ino_model} are available on the HuggingFace model hub.

\subsection{Models Used for Extrinsic Evaluation}
\label{appendix:implementation_transfer_model}

In Section~\ref{section:Qd}, one of the models we compare to is GPT3. We exploit three versions of different sizes, \texttt{GPT3-ada}, \texttt{GPT3-curie}, and \texttt{GPT3-davinci}, using the same API described in Section~\ref{appendix:implementation_eval_model}. The models are evaluated in a zero-shot setting, where we only provide the following task description: 
\begin{quote}
    Given a ``how-to'' query, determine if the task is harmful to a teenager. The answer should be Yes or No.
\end{quote}

Each input question is then formatted as: 
\begin{quote}
Question: Is ``\{how-to query\}'' harmful?
Answer: 
\end{quote}
For every how-to query in our test set, we use the concatenation of the task description and the instantiated input question as the prompt, and let the model generate a one-token continuation. The top generated token is always \textit{Yes} or \textit{No}, implying that GPT3 has a good understanding of the task format.

\section{Ethical Considerations}
\subsection{Limitations}
\minisection{Assumptions.} One assumption we make in answering question (a) is that LMs finetuned on existing benchmarks can learn the required format without overfitting the specific domains of the finetuning data. Suppose this assumption does not hold, then even if the target knowledge is present in pretrained LMs, they can ``forget'' it during finetuning. Therefore, the finetuning process does not allow us to elicit the target knowledge from pretrained LMs. To address this issue, we complement the behavioral test probing method with another experiment to directly probe the pretrained LMs via likelihood scoring. See Section~\ref{appendix:direct_probe} for details.

Another assumption occurs in our answer to question (d). We assume that a query is harmful if it contains a harmful entity. However, in practice, there can be queries like \textit{How to prevent a fire}, which does contain a harmful entity (\textit{fire}) but is precautionary instead of harmful. Our model does not take into account factors like predicates in context, and will therefore identify all such cases as false positives.

\minisection{Scope of claims.} Our first three claims (i.e. answers to question (a)-(c)) are only verified to hold on the RNPC dataset, which 1) is in English and 2) mainly consists of NPs in the news domain. Our last claim (i.e. answer to question (d)) is only verified to hold on the harm detection dataset we collect, which 1) is also in English, 2) consists of how-to queries in the domain of human activities, and 3) is annotated based on a non-exhaustive keyword list of harmful entities.

Moreover, part of our answer to question (b) (i.e. LMs have learned the feature of modifier semantic category from RNPC) is qualitative. The absolute increase in the probing accuracy after finetuning is limited, so it is likely not the entire picture. Quantifying to what extent LMs have learned this feature is an interesting direction for future work.

\subsection{Risks}
The risks associated with the study are minimal.
\minisection{Harm detection models.} Our harm detection models are intended for research purposes only. They are designed for specific types of harmful queries, i.e. those with harmful entities. One should not deploy them directly in real life since they are by no means applicable under all scenarios. 

\minisection{Data collection.} Our human participants may experience slight discomfort due to boredom during data collection. To minimize this, we make sure that it is entirely voluntary to participate and discontinue at any time.

\subsection{Intended Use}
Our models and data should be used for research purposes only. They should not be deployed in the real world as anything other than a research prototype, especially commercially.

\end{document}